# Comparative analysis of segmentation and generative models for fingerprint retrieval task

**Megh Patel[1*], Devarsh Patel[2], Sarthak Patel[3]**

[1]Department of Computer Science, Birla Institute of Technology and Science Pilani – KK Birla Goa Campus, Goa, India
[2]Department of Data Science, Indian Institute of Science Education and Research, Pune, India
[3]Department of Computer Science, Vellore Institute of Technology, Bhopal, India

*Corresponding author at
Email: meghpatel15@gmail.com

## ABSTRACT

Biometric Authentication like Fingerprints has become an integral part of the modern technology for authentication and verification of users. It is pervasive in more ways than most of us are aware of. However, these fingerprint images deteriorate in quality if the fingers are dirty, wet, injured or when sensors malfunction. Therefore, extricating the original fingerprint by removing the noise and inpainting it to restructure the image is crucial for its authentication. Hence, this paper proposes a deep learning approach to address these issues using Generative (GAN) and Segmentation models. Qualitative and Quantitative comparison has been done between pix2pixGAN and cycleGAN (generative models) as well as U-net (segmentation model). To train the model, we created our own dataset NFD [28] - Noisy Fingerprint Dataset meticulously with different backgrounds along with scratches in some images to make it more realistic and robust. In our research, the u-net model performed better than the GAN networks

## 1. Introduction

Human fingerprints are used for identification purposes due to its uniqueness, are difficult to alter and are immutable throughout a person's lifetime. This makes them a viable long-term biometrics option for identification and verification, which is evident as more fingerprint-based authentication methods are incorporated in day-to-day life tasks like unlocking smartphones, payments etc. Accuracy of fingerprint retrieval and verification systems is critical in forensic applications. However, recovery of fingerprints from surfaces like polished surfaces, metal, glass remains challenging [1-2].[1]

Factors like wetness, humidity, temperature or even non-uniform contact pressure with fingerprint sensor can severely affect fingerprint details, thus requiring a denoising method to remove fingerprint noise. Additionally, some regions of fingerprint can be missing due to fault in sensor, accidents, injuries or genetics [3]. To solve this, we need to inpaint patches of the image from neighboring regions.

Image inpainting and denoising are some preprocessing techniques that can be used to facilitate subsequent applications like identification and verification.

### 1.1. The objective of the paper

Recently, deep learning methods have been quite successful in dealing with aforementioned issues and are able to perform denoising and inpainting operations with high accuracy. In our paper, we frame the challenge of fingerprint retrieval from noisy photos as a denoising task. We explore generative and

---

[1] Unpublished working draft. Not for distribution

segmentation models and try to find which type is more suited for this task

## 1.2. Proposed novel work

Since fingerprints are extremely confidential pieces of information, none of the large fingerprint datasets are publicly available, which is why we built our own dataset NFD (Noisy Fingerprint Dataset) which contains image pairs of noisy and clean fingerprint images. We have used Anguli: Synthetic Fingerprint Generator [4] to generate distorted fingerprint samples and used a mix of images from Describable Textures Dataset (DTD) [5] as background to overlay noisy fingerprints on top. We hypothesize that using different kinds of image background will make the model robust to a wide variety of distortions.

Our aim is to retrieve fingerprints from a distorted image using a denoising and inpainting approach, where the foreground contains the noisy fingerprint and the background is a randomly selected texture. Generative models like GANs are excellent candidates to perform this task as they are frequently used in denoising and inpainting tasks. We have evaluated different GAN architectures on our NFD dataset and also explored the possibility to use segmentation model to segment fingerprint from background, hence phrasing it as segmentation task.

Our paper's three primary contributions are as follows: 1) We suggest NFD, a dataset of deformed and noisy fingerprint image pairings. 2) We propose a method to retrieve fingerprint from synthetically generated noisy fingerprint images. 3) We check the difference in performance of generative and segmentation models for this task

## 1.3. Paper Organization

The remaining portions of the essay are structured as follows: - Section 2 provides a summary of the previous comparable works. Section 3 describes our dataset, how it was created, and some of its key characteristics. The deployed deep learning models are explained in Section 4. Our training methodology is described in Section 5. It deals with experimental setup and data preprocessing. Section 6 contains the experiment results and comparisons between different models both qualitatively and quantitatively. The overview of our research and our predictions for future improvements can be found in Section 7.

## 2. Related work

Multiple approaches for enhancement of fingerprint have been suggested in literature but most of the preliminary efforts were based on simple image filtering algorithms like mean filtering, bilateral filtering, Gaussian filter, Laplacian filtering and many more. Some of the researches done on this domain has been listed below: -

1. Kuldeep et al. [6] demonstrated fingerprint denoising which makes use of a ridge orientation framework based on clustered sub dictionaries as well as sparse based denoising framework. This strategy groups areas with similar geometric features or dominating orientations, and then creates unique sub dictionaries for every single group. The gist of the work was to uplift the clarity of the ridge as well as valley patterns of the fingerprint image.

2. Mihir et al. [7] put forward a mechanism to denoise the fingerprint images with the help of unsupervised machine learning algorithms like boltzmann machines and multi-layered convolutional deep belief networks from converted grayscale images.

3. P. Venkadesh et al. [8] proposed an iterative approach for eradicating the noises in fingerprint and restructuring the same with the the help of 'iterative-rule based filter (IRF)'. This method consisted of five techniques, namely min-max based denoising, rule-based noise removal, mean and median computation.

4. Tang et al. [9] suggested a way to remove noises in the fingerprint with the help of FingerNet which is a Deep Convolutional Network (DCN) based model. For the extraction of fingerprint details in noisy ridge and valley patterns as well as complicated backgrounds, it leverages domain knowledge. The network divides the orientation field initially, then strengthens the latent fingerprint to gather details.

5. Antony et al. [10] approached the problem of fingerprint filtering using T2TRF filter, which is an algorithm specifically used for restoring images. The noisy pixels in the images were identified with the help of the Deep Convolutional Neural Networks and the detected noises were removed with the help of Rider Optimization Algorithm (ROA). They enhanced the pixels with the help of a type II fuzzy system. Metrics like Peak Signal to Noise Ratio (PSNR) and correlation coefficient were used to gauge the performance of their filter. They achieved the PSNR and correlation coefficient of 28.2467 db and 0.7504 respectively.

## 3. About our dataset

For fingerprint denoising and inpainting, we propose a new dataset NFD - Noisy Fingerprint Dataset. We explain how the fingerprints were created, how we added textured backdrops, and some salient features of our dataset in the parts that follow.

### 3.1. How the fingerprints were generated

Using Anguli: Synthetic Fingerprint Generator, we obtained pairs of noisy and ground-truth fingerprints. Anguli is an open-source C++ synthetic fingerprint generator designed with the intention of producing a large fingerprint database. Anguli is crucial to scholarly research on fingerprints and can be used to test large-scale government-deployed fingerprint recognition systems. Anguli creates noisy fingerprints by degrading ground truth fingerprints with a distortion model introducing blur, noise, translation, scrapes, and rotation. Our dataset consists of ground-truth and noisy fingerprint image pairings in the training, test, and validation sets as described in **Table 1**.

| Dataset type | No of Images |
|---|---|
| Train | 70,000 |
| Validation | 10,000 |
| Test | 20,000 |

*Table 1. NFD dataset*

### 3.2. How the background textures were added

We employ the Describable Textures Dataset (DTD), a growing collection of real-world textures that has been annotated with a set of human-centric characteristics that are motivated by the perceptual characteristics of textures, for our backdrop textures. We have 1250 texture files in total after filtering the photos. To add background textures to our noisy partial fingerprints, we randomly picked any of the texture files and then used the addWeighted() function of the OpenCV [11] library on these files. addWeighted() is a function that helps in alpha blending of the image. For 2 source images, $f_0(x)$ and $f_{1(x)}$, we can generate blended image using following relation:

$$g(x) = \alpha f_0(x) + (1 - \alpha)f_1(x)$$

By varying $\alpha$ from 0→1, we can vary the intensity of source images on output. We choose $\alpha = 0.45$ i.e, the weight of fingerprints is 0.45 while textures have 0.55 weight

### 3.3. Some advantageous features of our dataset

Some of the salient features of our dataset which enables it to perform well in large number of scenarios
1. The resolution of all images is 275×400 which provides sufficient raw data for a model
2. High variation in the collection is ensured by the large number of textures
3. The large variety in the background textures cover all real-life scenarios
4. Scratches are present in noisy fingerprints to simulate real life scratches and abrasions on our fingers due to accidents, injuries or genetics
5. Noisy images also have rotation (±10°) and translation (±10 pixels)

## 4. About models

For our research, to train the dataset we have used deep learning-based models like pix2pixGAN, cycleGAN and U-net for denoising the fingerprints. The mentioned models have been elaborated below.

### 4.1. pix2pixGAN

Generative Adversarial Networks (GANs) is an image-to-image translation neural network which

$$\min_G \max_D V(D, G)$$

$$V(D, G) = \mathbb{E}_{x \sim p_{data}(x)}[\log D(x)] + \mathbb{E}_{x \sim p_z(z)}[\log(1 - D(G(z)))]$$

*Equation 1. Mathematical description of generator and discriminator models of pix2pixGAN*

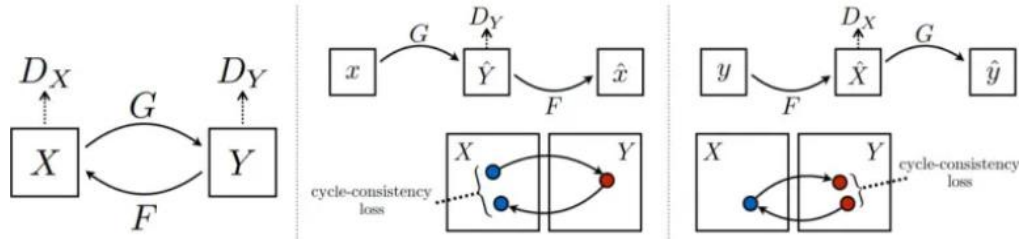

*Figure 1. cycleGAN architecture*

helps us to generate more refined and new sets of images from the existing images by impersonating them. GAN is bi-neural network model: Generative and Discriminative models which are independent and serve their own purpose. Generative model is responsible for generating new images from the base images and Discriminator as the name suggests is to recognize image as real or fake. By using a fixed size random noise as an input, the Generator Model creates new images. The Discriminator Model is then given the generated pictures [12-13].

The Generator's main objective is to deceive the Discriminator by creating images that mimic genuine images, making it more difficult for the Discriminator to distinguish between real and false images [14]. Generative and Discriminator model can be mathematically described as given in **Equation 1**.

G stands for Generator Model, D stands for Discriminator Model, z stands for fixed random noise, x stands for real image, G(z) stands for Generator Generated Image, pdata(x) stands for probability Distribution of Real Images, pz(z) stands for probability Distribution of Fake Images, D(G(z)) stands for Discriminator's Output for real image. That was the general overview of GAN.

Pix2PixGAN is a type of GAN which also helps in image-to-image translation as well as text to image translation. But, unlike the conventional GANs, The Pix2Pix model employs a conditional GAN (CGAN) as opposed to the traditional GAN model, which classifies pictures using a deep convolutional neural network. Instead of classifying the full input image as real or false, this deep convolutional neural network is intended to identify patches of the image. And as the name suggests "Pixel to Pixel", it means that it converts one pixel to another pixel but not for the whole image [15-16].

## 4.2. cycleGAN

CycleGAN is a special type of GAN which specializes in image-to-image translation models even when the images are not paired. pix2pixGAN and other GAN models requires the specific type of large dataset in which the related images are paired in the format of input and target image, but in the case of cycleGAN it is not necessary and it is smart to translate image even when the unrelated pairs of images are taken. cycleGAN has two generators and two discriminators as compared to pix2pixGAN. In a typical GAN model, the generator responsible for generating images and the discriminator is responsible for checking images as real or fake. But in the case of the CycleGAN, the second generator receives extra feedback from the first generator. This feedback makes sure that a picture produced by a generator is cycle consistent [17-18]. It means that if we translate image X to Y with the help of mapping (G) then G(x) to X should also return the same image. This network has 2 adversarial GAN and 1 cycle consistency loss. The architecture of CycleGAN is described in **Figure 1**.

Assume that there are two different image domains, X and Y. Since there are two domains, it has two mappings F: Y → X and G: X → Y along with DX and DY as two adversarial losses. DX will distinguish between images in the F(Y) and X domains. And DY will discriminate between images in the G(X) and Y domains. The cycle consistency loss function takes care of contradiction of learned mapping of F and G [19].

## 4.3. U-net

U-net is one of the semantic segmentation algorithms which was first used for Image Segmentation for Biomedical purposes in 2015. Image segmentation is a significantly more difficult operation that requires a sophisticated training architecture and a large amount of practice data. Image segmentation carries out two tasks: Localization and Classification. Localization means locating pixels of a particular object in a significantly larger image. And classification as the name suggests is to classify the object.

U-net as the name suggests is a U-shaped architecture consisting of encoders and decoders. The main function of this architecture is that it can localize and identify boundaries since every pixel is classified, ensuring that the input and output are of the same size [20-21]. U-net architecture can be understood with **Figure 2**.

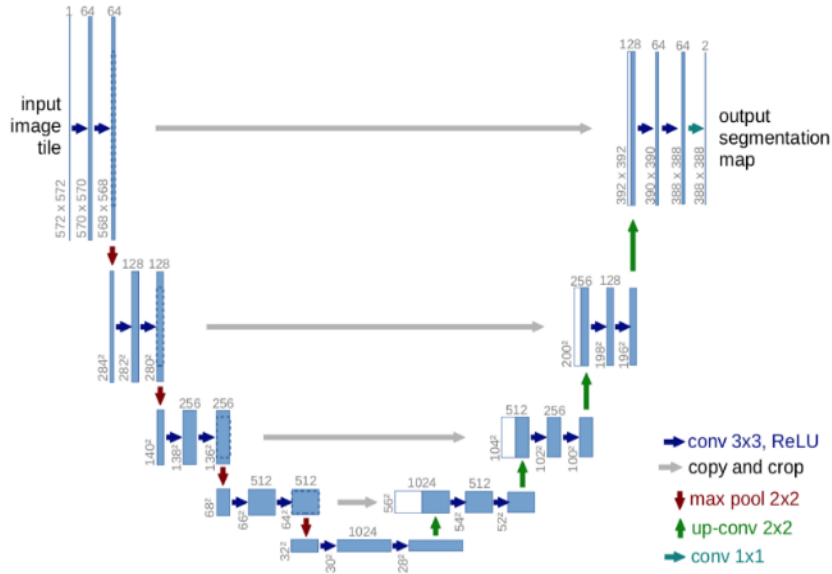

*Figure 2. U-net architecture*

The design is symmetrical and is divided into two main sections: the left section, known as the contracting path, is made up of the basic convolutional process, while the right section, known as the expansive path, is made up of transposed 2D convolutional layers. [22]. Transposed convolution layer is used for upsampling as it is basically used to enlarge the size of the images. The output image segmentation map is responsible for reshaping the image to meet the conditions of prediction requirements.

## 5. Methodology

In this section, we describe in detail the experiments conducted with generative and segmentation models on our NFD dataset. We also discuss in brief about how data was preprocessed before feeding to models.

### 5.1. Data preprocessing

Since our ground truth images can be represented as grayscale images, we converted all label images and set output channels to be 1. Images were cropped according to input size of models which is mentioned in Section 5.2.

### 5.2. Experimental setup

A Google colab notebook with an NVIDIA P100 GPU and 16GB of vRAM was used to train the network with different models. Pytorch [23] and Tensorflow [24] libraries were used to implement the complete design. Our dataset as well as code has been made available to the public.
We trained 2 generative models namely pix2pixGAN and cycleGAN and one segmentation model, UNet for comparative analysis.

#### 5.2.1. CycleGAN

CycleGAN was trained for 30 epochs over a week. Model was optimized using an Adam solver with a learning rate of 0.0002 and momentum parameters $\beta 1 = 0.5$, $\beta 2 = 0.999$. Weights were initialized from Gaussian distribution $\mathcal{N}(0, 0.02)$. Learning rate was linearly reduced to 0 every epoch after the 20th epoch. Batch size was chosen to be 64. Input image size of models was set to 256x256.

#### 5.2.2. pix2pixGAN

pix2pixGAN was optimized using minibatch SGD and using Adam solver with learning rate 0.0002 and momentum parameters $\beta 1 = 0.5$, $\beta 2 = 0.999$. Model was trained for a total of 35 epochs. We used the same linear learning rate decay method as CycleGAN with 15 decay epochs. Input images size was set to 256x256 with batch size of 6.

### 5.2.3. U-Net

We trained U-Net using input size of 256x256 and batch size of 32. Model was trained for 50 epochs with Adam optimizer having a learning rate of 0.0001. We used binary cross entropy for defining loss function.

## 6. Results and Discussion

The results of the above applied machine learning models have been compared both qualitatively as well as quantitatively with the help of performance metrics like Mean Square Error (MSE) as well as Peak Signal to Noise Ratio (PSNR). To further qualify the results, we have also evaluated Structural Similarity (SSIM) to affirm the perceptual quality of results. After metrics comparison, we are going to juxtapose all the models

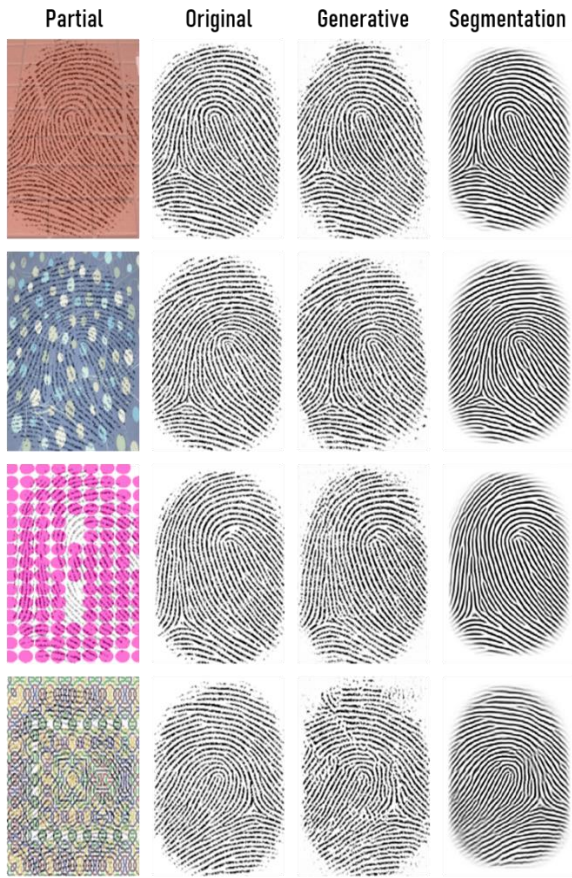

*Figure 3. Illustration showing the effects of fingerprint inpainting and denoising for varying distorted images. Left to right, in order: Partial or noisy fingerprints, Ground truth, best out of generative models' output, U-net output*

to analyze which class of models - generative or segmentation are superior for the task of fingerprint reconstruction

### 6.1. Qualitative analysis of models

Sample partial fingerprint images from the test set with their ground truths along with images predicted by our models are displayed in Fig 3. In each successive row, the distortion level gets progressively higher. Generative models were successful in extracting fingerprints even from the backgrounds like in Row 3 where it is hard even for naked eye to detect fingerprints. In case of heavily distorted images like in row 4, the models were not able to regenerate the partial image. Among GAN models, pix2pixGAN's results were superior as compared to cycleGAN despite it consuming a lot more computing power.

In comparison, U-net can successfully recover all prints and can withstand even substantial background noise (Row 4), although distortion can be seen in the outer parts of each image. As can be seen, we conclude that the U-net segmentation model outperforms both pix2pixGAN and cycleGAN in qualitative analysis.

### 6.2. Quantitative analysis of models

For quantitatively analyzing our models, we are going to use Image quality assessment metrics like MSE, PSNR which are frequently used because they are straightforward to calculate, have obvious physical implications, and are simple to apply mathematically in the context of optimization. However, they frequently lack the capacity to recognize visual quality and are not represented normally. To take care of that feature, we are also going to use Structured Similarity Indexing Method (SSIM). Before comparing models, let's take a brief insight on these metrics: -

#### 6.2.1. Mean Squared Error

MSE is used to calculate the combined square error of original and compressed images. The higher the value of MSE, the higher the error. MSE can be calculated as:

$$MSE = \frac{1}{mn} \sum_{0}^{m-1} \sum_{0}^{n-1} ||f(i,j) - g(i,j)||^2$$

For our practical needs, the mean squared error (MSE) enables us to contrast the "real" image pixels of our original image with those of our degraded image. The MSE is a measure of the "errors" between our real picture and our noisy image, and it is the average of those squares. The error is the difference between the values of the original and degraded images [25].

### 6.2.2. Peak Signal to Noise Ratio

Peak signal-to-noise ratio (PSNR) measures the relationship between a signal's maximum allowable value (power) and the power of distorted noise that reduces the signal's ability to be accurately represented. The PSNR mathematical formula is as follows:

$$PSNR = 20 log_{10}\left(\frac{MAX_f}{\sqrt{MSE}}\right)$$

Image compression quality is compared using the PSNR and the mean-square error (MSE). The PSNR provides insight into the peak error. The idea is that, higher the PSNR, degraded or noisy imaged has been better rebuilt to resemble the original image along with better reconstructive algorithm [26]. This would happen because we want to keep the MSE between images as close to the image's maximum signal value as possible.

### 6.2.3. Structural Similarity Index

A perceptual metric called the Structural Similarity Index (SSIM) measures how much image quality is lost during processing like data compression or during data transfer. Image quality assessment metrics like MSE and PSNR rely on evaluating the quality of an image and focus on calculating the differences between a reference image and a test image but SSIM replicates the human behavior of extracting differences between the reference and input image. SSIM metrics extricates three significant features from the image: - luminance, contrast and structure. SSIM applies aforementioned measurements regionally that is taking the mean of the overall small sections of the image rather than across the entire image at once. SSIM can be mathematically expressed as:

$$SSIM(x, y) = \frac{(2\mu_x\mu_y + c_1)(2\sigma_{xy} + c_2)}{(\mu_x^2 + \mu_y^2 + c_1)(\sigma_x^2 + \sigma_y^2 + c_2)}$$

The performance and computation of the deep learning models has been done quantitatively and has been depicted in **table 2** [27]. It can be concluded from the **table 2** that U-net performed superiorly than pix2pixGAN and cycleGAN in all the metrics. Lower the Mean Square Error (MSE) better the performance and U-net has lowest of all. None of the Generative Models provided better results than the segmentation model (U-net). But among the GAN models, pix2pixGAN performed slightly better than cycleGAN, especially in PSNR where the former exceeded the latter in significant numbers.

| Model | MSE | PSNR | SSIM |
|---|---|---|---|
| **pix2pixGAN** | 0.1109 | 9.6470 | 0.4627 |
| **cycleGAN** | 0.1362 | 7.9381 | 0.4055 |
| **U-Net** | 0.0466 | 13.4016 | 0.7714 |

*Table 2. Quantitative results*

## 7. Conclusion

In this paper, *we* propose a new dataset NFD - Noisy Fingerprint Dataset. Our dataset consists of pairs of synthetically generated ground-truth and noisy fingerprints. Noisy prints were obtained by degrading ground truth images by adding blur, noise, translation, rotation along with different textured backgrounds and scratches to make it more realistic and robust.

Using NFD, this paper explores a deep learning approach to extricate the original fingerprint from noisy images using Generative and Segmentation models. Qualitative and Quantitative comparison has been done between pix2pixGAN and cycleGAN (generative models) as well as U-net (segmentation model). In our research, the u-net model performed better than the GAN networks suggesting that this task is more suited to segmentation models